\pdfoutput=1

\documentclass[11pt]{article}

\usepackage{ACL2023}

\usepackage{times}
\usepackage{latexsym}

\usepackage[T1]{fontenc}

\usepackage[utf8]{inputenc}

\usepackage{microtype}

\usepackage{inconsolata}

\usepackage{graphicx}
\usepackage{subcaption}
\usepackage{booktabs}
\usepackage{adjustbox}
\usepackage{array}
\usepackage{colortbl}
\usepackage{multirow}
\usepackage{stfloats} 
\usepackage{censor}
\usepackage{makecell}
\usepackage{mathtools}

\newcolumntype{R}[2]{%
    >{\adjustbox{angle=#1,lap=\width-(#2)}\bgroup}%
    l%
    <{\egroup}%
}
\newcommand*\rot{\multicolumn{1}{R{35}{1em}}}

\title{A Better Way to Do Masked Language Model Scoring}

\author{Carina Kauf \\
      Massachusetts Institute of Technology \\
      \href{mailto:ckauf@mit.edu}{\tt ckauf@mit.edu}\And
      Anna A. Ivanova \\
      Massachusetts Institute of Technology \\
      \href{mailto:annaiv@mit.edu}{\tt annaiv@mit.edu}}

\begin{document}
\maketitle
\begin{abstract}
Estimating the log-likelihood of a given sentence under an autoregressive language model is straightforward: one can simply apply the chain rule and sum the log-likelihood values for each successive token. However, for masked language models (MLMs), there is no direct way to estimate the log-likelihood of a sentence. To address this issue, \citet{salazar2020masked} propose to estimate sentence pseudo-log-likelihood (PLL) scores, computed by successively masking each sentence token, retrieving its score using the rest of the sentence as context, and summing the resulting values. Here, we demonstrate that the original PLL method yields inflated scores for out-of-vocabulary words and propose an adapted metric, in which we mask not only the target token, but also all within-word tokens to the right of the target. We show that our adapted metric ({\tt PLL-word-l2r}) outperforms both the original PLL metric and a PLL metric in which all within-word tokens are masked. In particular, it better satisfies theoretical desiderata and better correlates with scores from autoregressive models. Finally, we show that the choice of metric affects even tightly controlled, minimal pair evaluation benchmarks (such as BLiMP), underscoring the importance of selecting an appropriate scoring metric for evaluating MLM properties.\footnote{Our results and code are available at \url{https://github.com/carina-kauf/better-mlm-scoring}.}

\end{abstract}

\section{Introduction}

Most state-of-the-art transformer-based large language models (LLMs) fall into two classes: unidirectional (or autoregressive) models, where each token is generated based on its left context (e.g., GPT models; \citealp{radford2019language}), and bidirectional models, where a token is predicted from both left and right context tokens, some of which may be masked (e.g., BERT; \citealp{devlin2018bert}). Often, it is beneficial to compare these models' performance on controlled sentence generation benchmarks. Whereas unidirectional architectures offer a natural way of calculating sentence log-likelihood (summing the log-likelihood scores of each sentence token given its left context), there is no direct way of estimating sentence log-likelihood for a bidirectional model.

So far, the best available method to score a sentence under a bidirectional LLM has been the pseudo-log-likelihood (PLL) scoring approach described by \citet{salazar2020masked} (and initially used by \citealp{shin2019effective, wang-cho-2019-bert}). The PLL of a sentence is calculated as the sum of PLL scores for each token given all other sentence tokens, thus providing a comparable metric to unidirectional models' log-likelihood (LL) sentence scoring. The PLL metric is extremely popular; it is used extensively in LLM studies tackling topics as diverse as effects of training data \citep{sinha-etal-2021-masked, zhang-etal-2021-need}, model fluency \citep{laban-etal-2021-keep}, syntactic and conceptual knowledge \citep{sinclair2022structural, bhatia2022transformer}, social biases \citep{nangia-etal-2020-crows}, and others. Some of these studies have already accrued dozens of citations.

\begin{figure}[tb!]
   \centering
   \includegraphics[width=1\columnwidth]
   {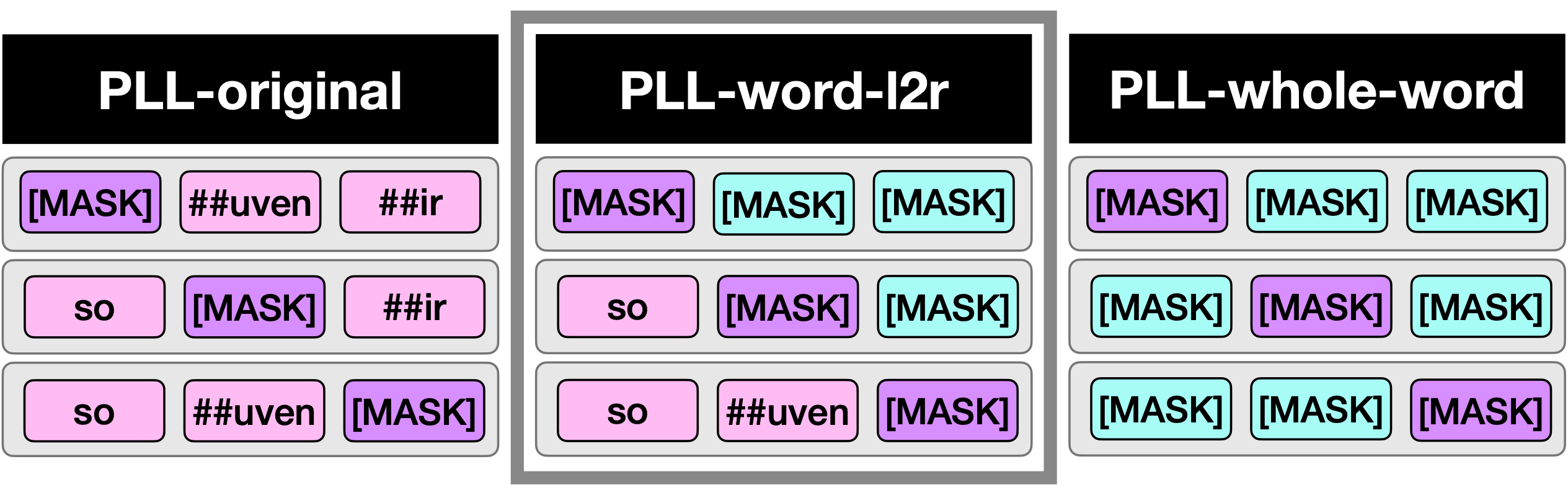}
           \caption{Three different ways to compute the PLL score of a multi-token word (e.g., {\tt souvenir}) during masked language modeling. \emph{Purple}: target token, \emph{pink}: within-word tokens that are available during inference, \emph{turquoise}: within-word tokens that are masked during inference. Sentence tokens that do not belong to the current word are always available during inference.
   }
   \label{fig:conceptual}
\end{figure}

\begin{figure*}[t!]
    \centering
    \includegraphics[width=\textwidth]
    {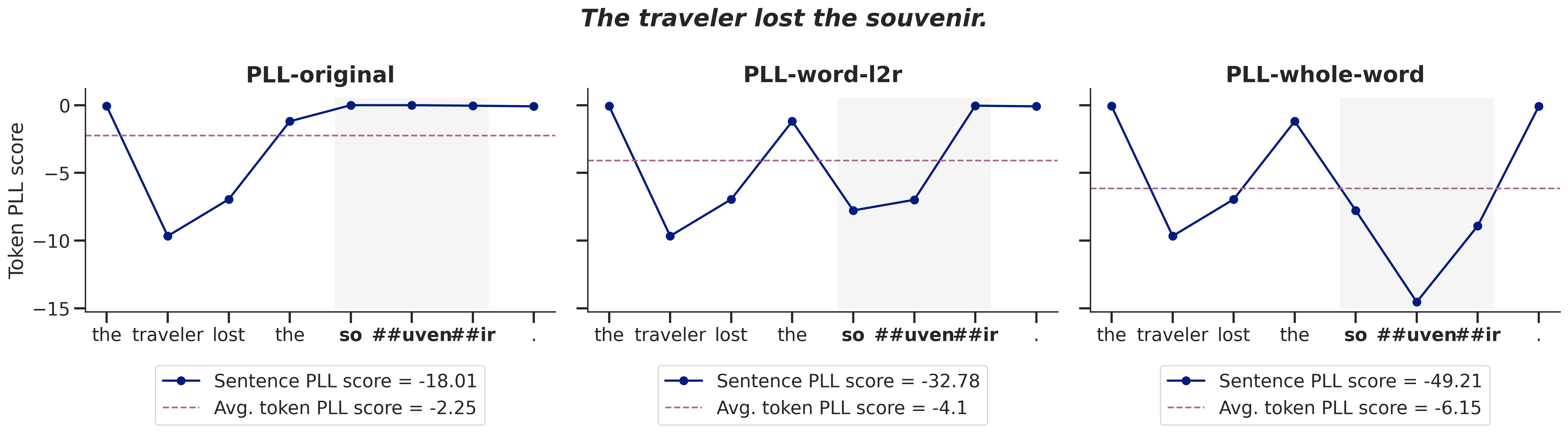}
    \caption{The {\tt PLL-original} metric inflates scores of multi-token words, such as \emph{souvenir}; the adjusted metrics, {\tt PLL-word-l2r} and {\tt PLL-whole-word}, mitigate this issue. Example generated using the {\tt bert-base-cased} model.}
    \label{fig:motivation-for-adaptation}
\end{figure*}

Here, we show that the metric proposed by \citeauthor{salazar2020masked} ({\tt PLL-original}) has important shortcomings that limit its utility. Specifically, {\tt PLL-original} overestimates the PLL of out-of-vocabulary (OOV) words, which LLM tokenizers split into multiple tokens. As a result, {\tt PLL-original} scores fail on several theoretically desired property tests: a robust inverse relationship between sentence length and sentence PLL (Section \ref{sec:sentlength}), a robust positive correlation between a word's frequency and its PLL score (\ref{sec:wordfreq}), and a positive correlation between unidirectional and bidirectional model scores for the same sentences (Section \ref{sec:betweenmodelcorrelation}).
To remedy these issues, we propose an adjusted PLL metric, {\tt PLL-word-l2r} (l2r: left-to-right), which estimates token PLL when future within-word tokens are also masked (Figure \ref{fig:conceptual}). We show that the {\tt PLL-word-l2r} metric outperforms both {\tt PLL-original} and alternative PLL-based metrics. We therefore recommend to use the {\tt PLL-word-l2r} metric when estimating sentence PLL under a bidirectional LLM.

\section{Motivation: score inflation for multi-token words}

The {\tt PLL-original} metric grossly overestimates the probability of OOV lexical items, such as \emph{souvenir} (Figure \ref{fig:motivation-for-adaptation}). This is because OOV words are tokenized into subword tokens (e.g., \emph{so \#\#uven \#\#ir}), and each subword token is predicted using the token's bidirectional context, which crucially includes the remaining tokens that make up the OOV word. Thus, even though the OOV word itself may be surprising given the sentence context, the individual parts of the OOV word are not surprising to a bidirectional model given a sentence context that includes all other subtokens of that word (e.g., it is easy to predict \emph{so} given \emph{\#\#uven \#\#ir}; see Appendix \ref{app:examples} for additional examples).

To mitigate this bias, we adjust the PLL sentence scoring algorithm such that the model cannot access future within-word tokens ({\tt PLL-word-l2r}) or any within-word tokens ({\tt PLL-whole-word}) when predicting the target.

Below, we conduct a rigorous investigation of our modified metrics to determine whether this intuitive benefit holds quantitatively.

\begin{figure*}[ht]
    \centering
    \includegraphics[width=\textwidth]{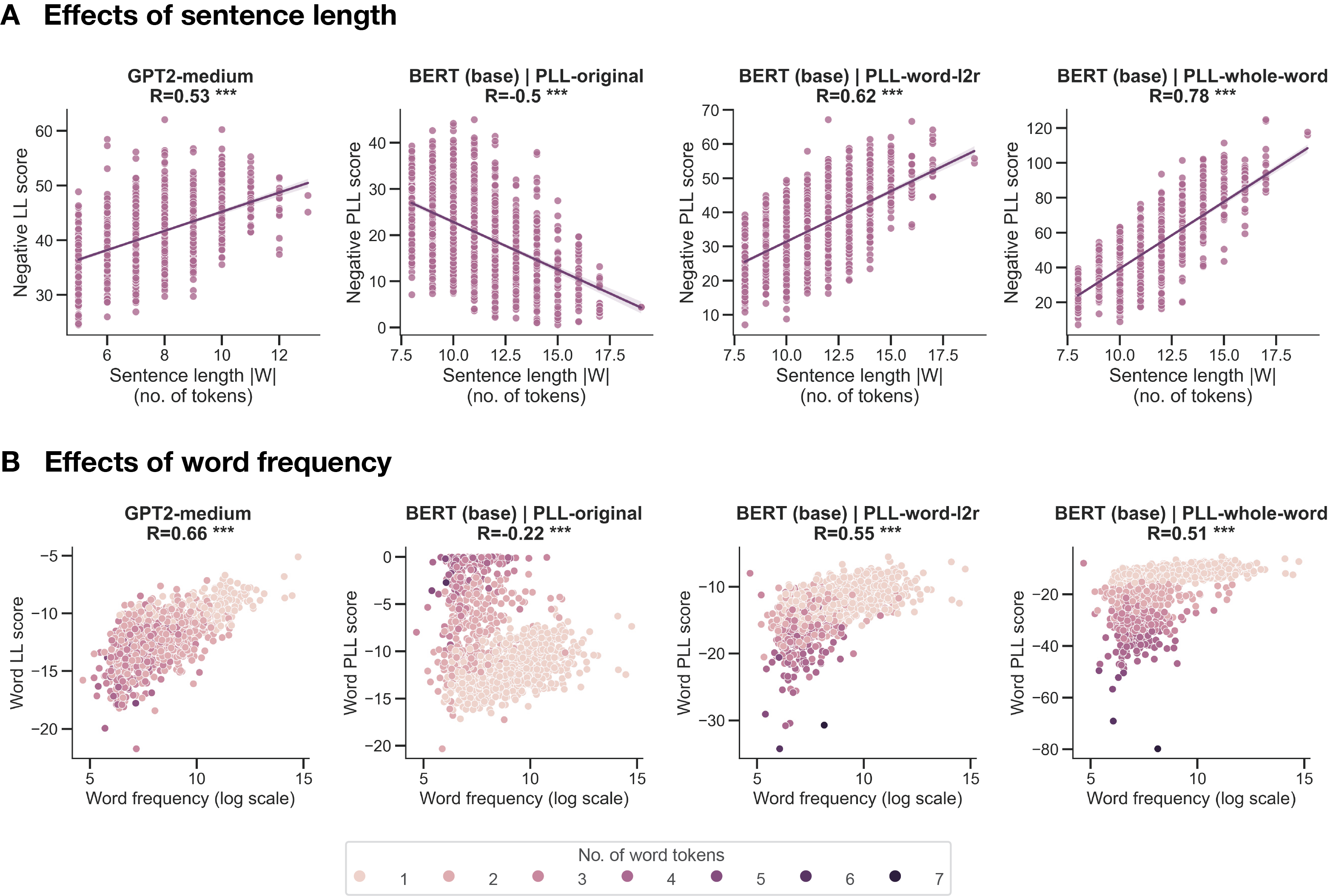}
    \caption{Out of all PLL metrics, {\tt PLL-word-l2r} best satisfies theoretical desiderata: \textbf{(A)} an inverse relationship between negative sentence PLL (a measure of model surprisal) and sentence length and \textbf{(B)} a positive correlation between word PLL and word log frequency. In (A), each dot is a sentence; in (B), each dot is a unique word from the dataset. Here and elsewhere, reported correlations are Pearson correlations.} 
    \label{fig:properties}
\end{figure*}

\section{Methods}
For our analysis, we adapt the {\tt scorer} module of the {\tt minicons} library \cite{misra2022minicons}, an open-source wrapper library around HuggingFace {\tt transformers} \cite{wolf2020transformers} that enables efficient extraction of word- and sentence-level probabilities from LLMs. The MLM scoring procedure of the {\tt minicons} library follows the procedure originally proposed by \citet{salazar2020masked}. For details on sentence preprocessing, see Appendix \ref{app:textprep}.

\subsection{PLL metrics}
{\tt PLL-original}. In this metric, each sentence token $s_t$ of a sentence $S$ with $n$ tokens is consecutively replaced with a {\tt [MASK]} and is predicted using all past and future tokens, irrespective of whether the context tokens belong to the same or a different word than the target token. Thus, inference is conditioned on the context $S_{\setminus t} := (s_1, \dots ,s_{t-1},s_{t+1}, \dots ,s_n)$. The final sentence score is obtained as the sum of the log probabilities of each sentence token given its context:
\begin{equation}
    \mathrm{PLL}_{\mathrm{orig}}(S) := \sum_{t=1}^{n} \mathrm{log}~P_{\mathrm{MLM}}(s_t~|~S_{\setminus t})
\end{equation}

\noindent
{\tt PLL-word-l2r}. In this metric, a {\tt [MASK]} is placed not only over the current target token (now: $s_{w_t}$), but also over all future sentence tokens that belong to the same word $s_w$ as the target. Inference is then conditioned on a context that includes all preceding sentence tokens (including those belonging to the current word) and all sentence tokens from future words. The final score of a sentence $S$ is obtained as the sum of the log probabilities of each of the $|w|$ tokens in each of the $|S|$ words:

\begin{equation}
    \mathrm{PLL}_{\mathrm{l2r}}(S) := \sum_{w=1}^{|S|} \sum_{t=1}^{|w|} \mathrm{log}~P_{\mathrm{MLM}}(s_{w_t}~|~S_{\setminus s_{w_{t' \geq t}}})
\end{equation}

\noindent
{\tt PLL-whole-word}. This metric is similar to {\tt PLL-word-l2r} and differs from it only in that a {\tt [MASK]} is placed over \emph{all} sentence tokens that belong to the same word $s_w$ as the target (both preceding and future). Inference is then conditioned on a context that includes all sentence tokens except those belonging to the current word. The final score of a sentence $S$ is obtained as the sum of the log probabilities of each of the $|w|$ tokens in each of the $|S|$ words in $S$ given the token's context:

\begin{equation}
    \mathrm{PLL}_{\mathrm{ww}}(S) := \sum_{w=1}^{|S|} \sum_{t=1}^{|w|} \mathrm{log}~P_{\mathrm{MLM}}(s_{w_t}~|~S_{\setminus s_w})
\end{equation}

In Appendix \ref{app:sentence-l2r}, we also report results for a PLL metric where not only future within-word tokens, but \textit{all} sentence tokens to the right of the target context are masked ({\tt PLL-sentence-l2r}). Although this method is most similar to autoregressive LL scoring, {\tt sentence-l2r} masking for BERT is known to produce poor quality generations \citep{wang-cho-2019-bert}; we therefore refrain from including this metric in the main text.

\begin{figure*}[ht]
    \centering
    \includegraphics[width=\textwidth]{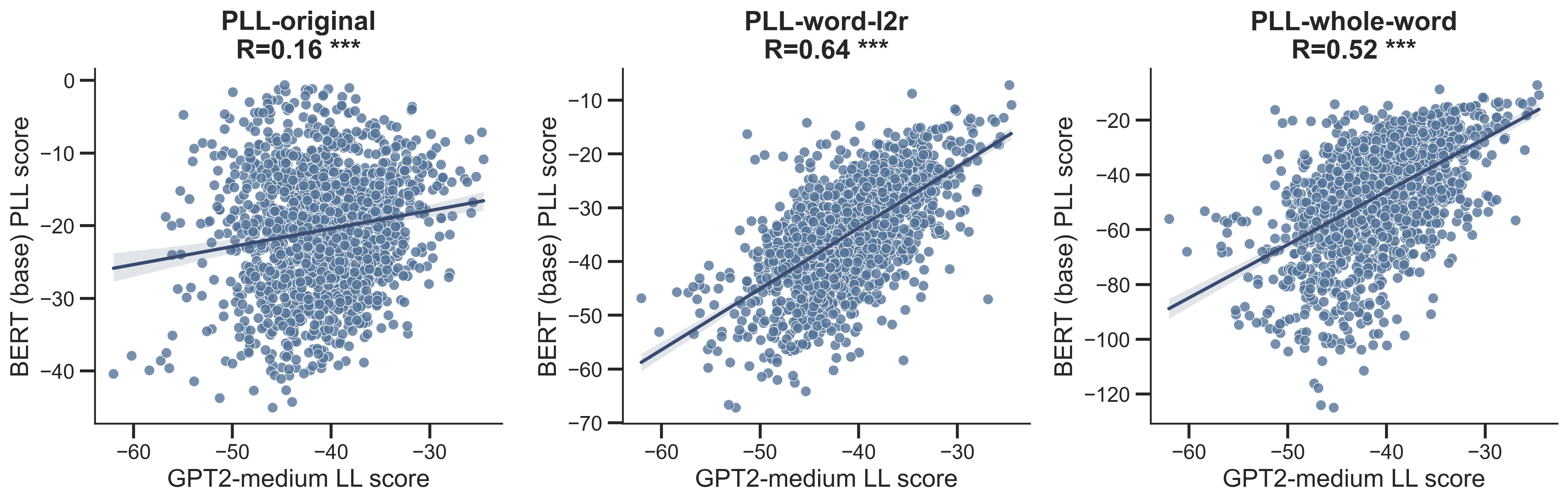}
    \caption{Correlation between bidirectional model PLL scores and unidirectional model LL scores. Each dot is a sentence.} 
    \label{fig:gpt-vs-bert}
\end{figure*}

\subsection{Models}

We report results for {\tt bert-base-cased} (and {\tt gpt2-medium} for comparison) unless stated otherwise. Results for larger models are provided in Appendices \ref{app:sentence-length}-\ref{app:gpt-correlation}.

\subsection{Datasets}
For our main analyses, we use the EventsAdapt dataset \cite[based on \citealp{fedorenko2020lack}]{kauf2022event}. It contains a curated set of 782 syntactically simple sentence pairs that describe plausible or implausible agent-patient interactions in active or passive voice (e.g., \emph{The traveler lost the souvenir}). Sentences in this dataset are 5-7 words long (mean: $6.1$, std: $1.05$), with an average word log frequency of 10.95. We use this dataset because it contains a high number of OOV words (19.6\% for BERT and 40.3\% for GPT-2; see also Appendix \ref{app:OOVratio}). In Appendices \ref{app:sentence-length}-\ref{app:gpt-correlation}, we show that our results generalize to two larger and more diverse corpora: the Brown corpus \cite{francis1979brown} and the reference sentence set from the LibriSpeech corpus \cite{panayotov2015librispeech}. We also apply our PLL metrics to score the sentences in the Benchmark of Linguistic Minimal Pairs (BLiMP) \cite{warstadt2020blimp}, a challenge set of 67k sentence pairs which target specific aspects of linguistic knowledge.

\section{Evaluating PLL metric properties}

\subsection{Effects of sentence length} \label{sec:sentlength}

Like \citet{salazar2020masked}, we expect that models should, on average, assign lower probability to longer sentences. Thus, negative PLL (which reflects model surprisal) should be positively correlated with sentence length. However, the {\tt PLL-original} metric violates this expectation in our test sentence set, which shows a negative correlation between the number of tokens and negative PLL. In contrast, {\tt PLL-word-l2r}  and {\tt PLL-whole-word} metrics exhibit a positive correlation between the number of sentence tokens and negative PLL, just as the negative LL scores for a unidirectional model, GPT2-medium (Figure \ref{fig:properties}A).

\subsection{Effects of word frequency} \label{sec:wordfreq}

An appropriate (P)LL metric should reflect the fact that LLMs are sensitive to distributional patterns in training text corpora. In particular, we expect more frequent words to have higher (P)LL scores in the absence of contextual effects. This is indeed the case for GPT2-medium; however, the score inflation for multi-token words means that the {\tt PLL-original} metric grossly overestimates the scores for low-frequency words (Figure \ref{fig:properties}B). {\tt PLL-word-l2r} scores restore this relationship: their correlation with word frequency is much higher than for {\tt PLL-original}. {\tt PLL-whole-word} also performs well, although its correlation with word frequency is lower than for {\tt PLL-word-l2r}, suggesting that it excessively penalizes OOV words.

\section{Correlation with GPT-2 scores} \label{sec:betweenmodelcorrelation}

We expect that PLL scores for bidirectional models should be at least somewhat consistent with LL scores for unidirectional models: both metrics are designed to serve are a proxy for sentence probability. Here, we show that the GPT-2/BERT score correlation for the {\tt PLL-original} metric is very low, whereas correlation scores for {\tt PLL-word-l2r} and {\tt PLL-whole-word} are much higher (Figure \ref{fig:gpt-vs-bert}), indicating the validity of this metric for cross-model comparison. As in Section \ref{sec:wordfreq}, {\tt PLL-word-l2r} slightly outperforms {\tt PLL-whole-word}, likely because it does not penalize OOV words as severely. 

See Appendices \ref{app:sentence-length}-\ref{app:gpt-correlation} for evidence that all three trends hold for larger models and for other datasets (although the effects in other datasets are attenuated due to a lower OOV ratio).

\begin{table}[!ht]
\centering
  \footnotesize
\begin{tabular}{ll>{\columncolor[gray]{0.95}}c}
\textbf{Model} & \textbf{Metric} &  \textbf{Overall score} \\
\toprule
\multirow{3}{*}{BERT (base)} & PLL-original       &     84.2 \\
 & PLL-word-l2r       &     \textbf{84.7}  \\
 & PLL-whole-word     &     83.1 \\
\midrule
\multirow{3}{*}{BERT (large)} & PLL-original      &     84.8  \\
 & PLL-word-l2r      &     \textbf{85.0} \\
 & PLL-whole-word    &     82.6 \\
\midrule
\multirow{3}{*}{RoBERTa (base)} & PLL-original    &     85.4 \\
 & PLL-word-l2r    &     \textbf{86.7} \\
 & PLL-whole-word  &     85.4 \\
\midrule
\multirow{3}{*}{RoBERTa (large)} & PLL-original   &     86.5 \\
 & PLL-word-l2r   &     \textbf{87.5} \\
 & PLL-whole-word &     85.9 \\
\bottomrule
\end{tabular}
\caption{\normalsize Bidirectional model performance on the BLiMP benchmark using different PLL metrics.}
\label{table:BLiMP}
\end{table}

\section{Effects on benchmarking}

Here, we show that the choice of PLL metric affects benchmarking results for a popular, highly controlled, minimal pair linguistic benchmark: BLiMP. Despite the fact that the comparisons are highly controlled, different metrics yield different BLiMP scores. For all four tested models, {\tt PLL-word-l2r} achieves the best overall BLiMP score (Table \ref{table:BLiMP}). See Appendix \ref{app:blimp} for detailed scores.

\section{Conclusion}

We have shown that {\tt PLL-word-l2r} is the preferred metric for evaluating sentence PLL under a masked language model, such as BERT. Although the results from studies using the {\tt PLL-original} metric can still be informative, they become harder to interpret if the proportion of OOV words in their test set is high. Therefore, we recommend using {\tt PLL-word-l2r} in future works.

\section*{Limitations}

The proposed {\tt PLL-word-l2r} metric has the same practical limitations as previous LL/PLL approaches. Most importantly, these scores can be influenced by many superfluous factors, such as the number of available synonyms (\emph{computer} vs. \emph{laptop}; \citealp{holtzman-etal-2021-surface}). We therefore expect our method to be most useful in highly controlled minimal pair or multiple choice setups.

Even more accurate metrics may emerge in the future. For instance, our approach pre-specifies the number of tokens in a word, thus limiting the space of possible alternatives. Future approaches might investigate a way to normalize the PLL score distribution over words with a varying number of tokens. Further, it would be interesting to attempt to estimate the joint probability of all tokens in a word instead of predicting them left-to-right (as in {\tt PLL-word-l2r}) or without any other within-word contextual information (as in {\tt PLL-whole-word}).

Finally, we test our approach on English text corpora; our results might not generalize to agglutinative languages (due to a high number of tokens per word and, therefore, increased uncertainty) and are of less relevance to isolating languages (where, if enough training data are available, most word-level items can be represented as single tokens).

\section*{Ethics Statement}
 
In our proposed metric, word tokens are masked from left to right following the writing tradition in English; however, for speakers of languages such as Arabic, a ``right to left'' notation would be more intuitive. Note, however, that this is primarily a denotational difference that does not affect the score itself (LLMs do not discriminate left and right, only beginning and end). We do not anticipate any specific harms that would be intrinsically associated with the techniques described in this paper.

\section*{Acknowledgements}
We thank Jacob Andreas, Evan Hernandez, and the anonymous ACL reviewers for their insightful feedback. CK was supported by the K. Lisa Yang Integrative Computational Neuroscience (ICoN) Center at MIT. AI was supported by MIT Quest for Intelligence.

\bibliography{custom}
\bibliographystyle{acl_natbib}


\appendix
\section*{Appendix} \label{sec:appendix}

\section{Additional examples of score inflation} \label{app:examples}

\noindent%
\begin{minipage}{\linewidth}
\makebox[\linewidth]{
  \includegraphics[width=\textwidth]{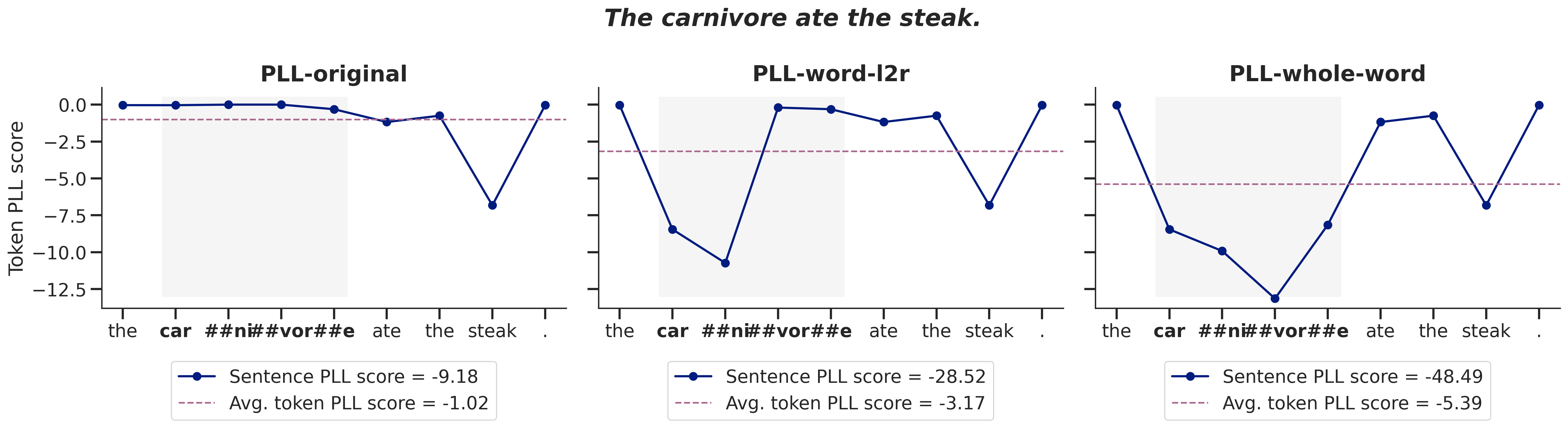}}
\captionof{figure}{The {\tt PLL-original} metric inflates the score of the word \emph{carnivore}. {\tt PLL-word-l2r} mitigate this issue, whereas {\tt PLL-whole-word} overly penalizes the word. Model: {\tt bert-base-cased}.}
\label{fig:motivation-carnivore}
\end{minipage}

\noindent%
\begin{minipage}{\linewidth}
\makebox[\linewidth]{
  \includegraphics[width=\textwidth]{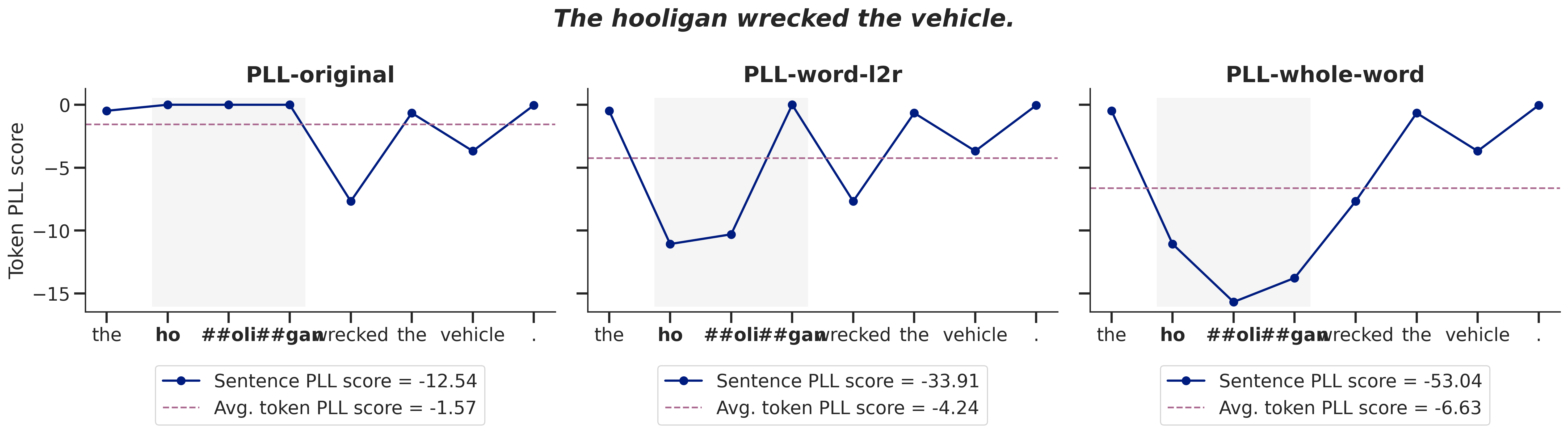}}
\captionof{figure}{The {\tt PLL-original} metric inflates the score of the word \emph{hooligan}. {\tt PLL-word-l2r} mitigate this issue, whereas {\tt PLL-whole-word} overly penalizes the word. Model: {\tt bert-base-cased}.}
\label{fig:motivation-hooligan}
\end{minipage}

\section{Text preprocessing for (P)LL computation} \label{app:textprep}

The {\tt minicons} library borrows the MLM preprocessing algorithm from \citet{salazar2020masked}: {\tt [CLS]} and {\tt [SEP]} tokens are prepended and appended to the text, respectively, and are not masked during PLL computation. For CLMs, we minimally adjust the {\tt minicons} scorer library default and necessarily prepend the beginning of sentence token, {\tt <|endoftext|>}, to the text, which enables us to get a probability for the first actual sentence token (see also the {\tt lm-zoo} library; \citealp{gauthier2020syntaxgym}). The (P)LLs of all special tokens are not counted toward the final sentence/word score.

When calculating the (P)LL score of individual words (to estimate word frequency effects), we place them in a neutral context \emph{My word is \_}. To ensure that the same pattern of results holds across multiple neutral contexts, we additionally test the context \emph{I opened the dictionary and randomly picked a word. It was \_}, as well as a no-context setup. These additional results are reported in Appendix \ref{app:frequency-contexts}.

Word frequency was operationalized as the log of the number of occurrences of the word in the 2012 Google NGram corpus. Laplace smoothing was applied prior to taking the logarithm.

\section{Quantification of out-of-vocabulary words per dataset} \label{app:OOVratio}

\begin{table}[!ht]
\centering
  \footnotesize
\begin{tabular}{ll>{\columncolor[gray]{0.95}}c}
\textbf{Dataset} & \textbf{Model class} & \textbf{OOV ratio} \\
\toprule
\multirow{3}{*}{EventsAdapt} & BERT       &     19.6\% \\
& RoBERTa       &     40.3\% \\
& GPT       &     40.3\% \\
\midrule
\multirow{3}{*}{LibriSpeech} & BERT     &     8\%  \\
& RoBERTa       &     24.3\% \\
& GPT       &     24.3\% \\
\midrule
\multirow{3}{*}{Brown} & BERT    &     8\% \\
& RoBERTa       &     25\% \\
& GPT       &     25\% \\
\bottomrule
\end{tabular}
\caption{\normalsize The out-of-vocabulary (OOV) ratio per dataset, quantified as the number of words split into at least two tokens by a given model's tokenizer divided by the total number of words in the dataset. GPT and RoBERTa models use byte-level Byte-Pair-Encoding tokenizers \cite{radford2019language, liu2019roberta}; BERT models use WordPiece tokenization \cite{devlin2018bert}.}
\label{table:OOV}
\end{table}


\section{Effects of sentence length} \label{app:sentence-length}

\subsection{Larger LLM versions}

\noindent%
\begin{minipage}{\linewidth}
\makebox[\linewidth]{
  \includegraphics[width=\textwidth]{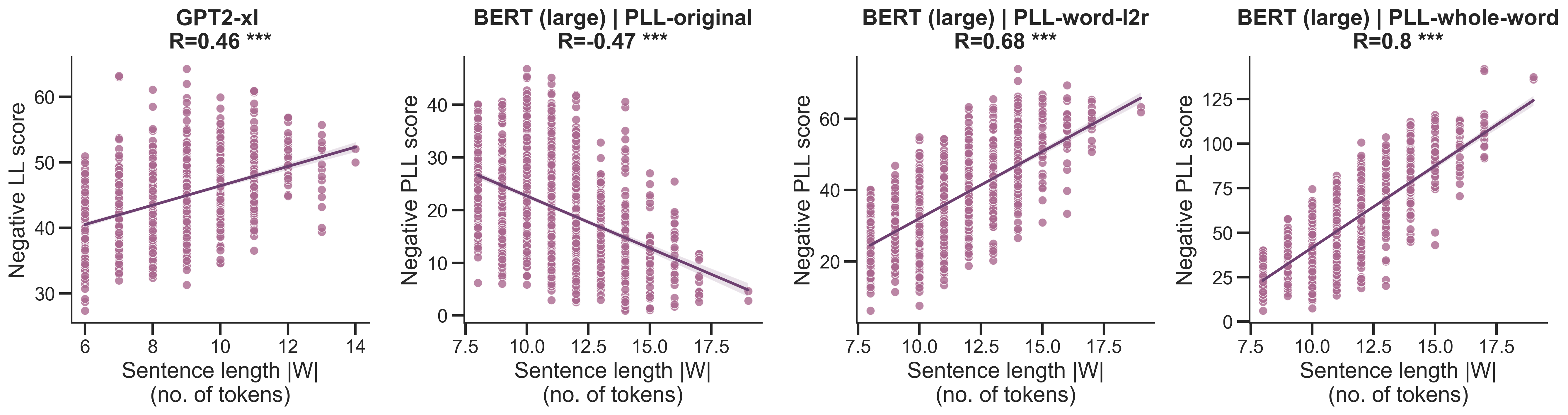}}
\captionof{figure}{Sentence length effects for {\tt gpt2-xl} and {\tt bert-large-cased} on the EventsAdapt corpus.} 
\label{fig:sent-length-vs-PLL-1}
\end{minipage}

\subsection{Larger datasets}

\noindent%
\begin{minipage}{\linewidth}
\makebox[\linewidth]{
  \includegraphics[width=\textwidth]{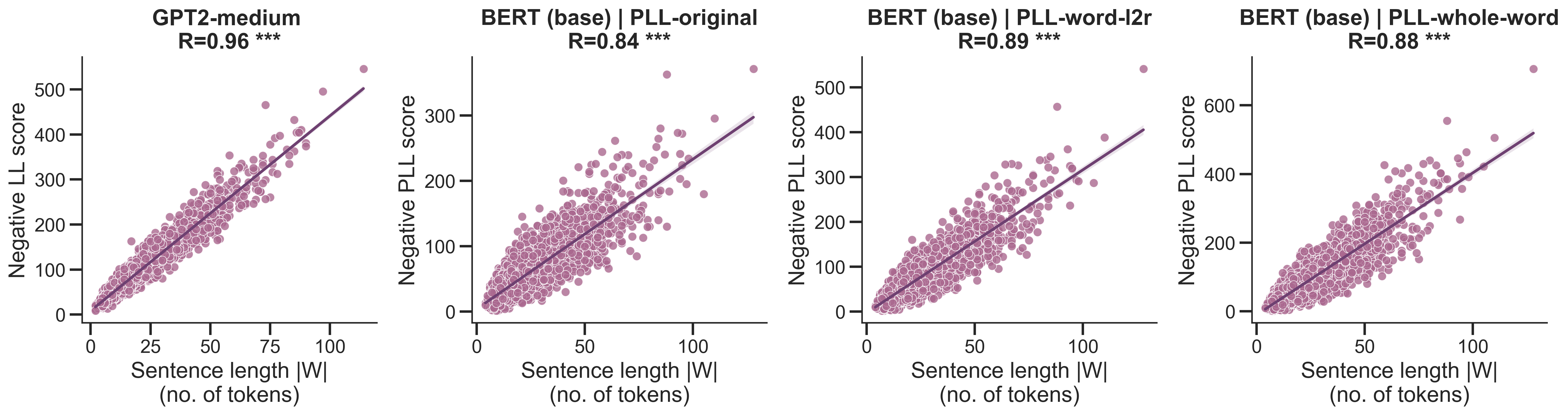}}
\captionof{figure}{Sentence length effects for {\tt gpt2-medium} and {\tt bert-base-cased} on the LibriSpeech corpus.\newline}
\label{fig:sent-length-vs-PLL-2}
\end{minipage}

\noindent%
\begin{minipage}{\linewidth}
\makebox[\linewidth]{
  \includegraphics[width=\textwidth]{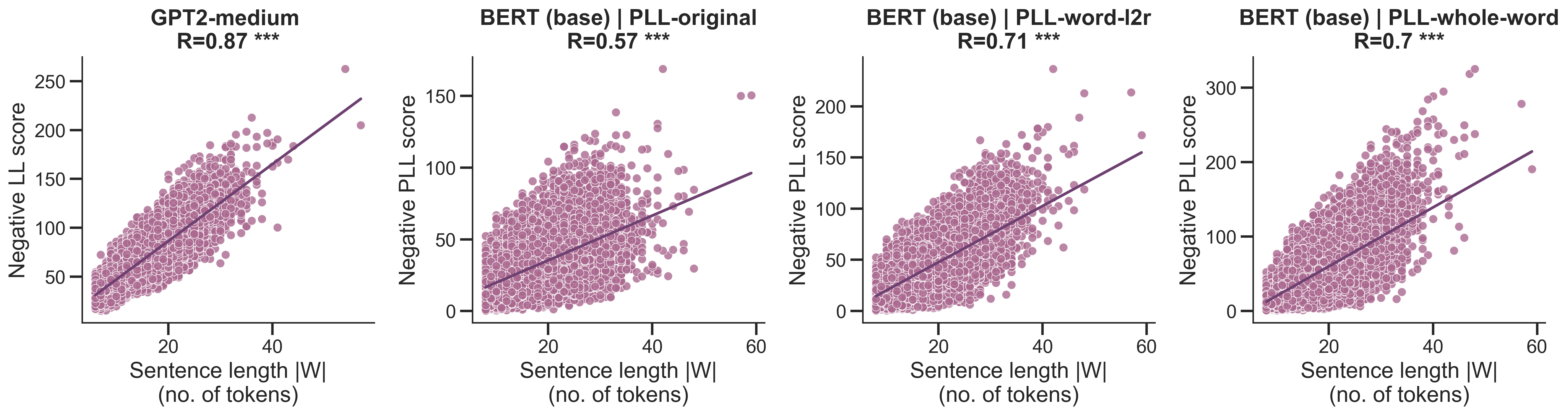}}
\captionof{figure}{Sentence length effects for {\tt gpt2-medium} and {\tt bert-base-cased} on the Brown corpus.}
\label{fig:sent-length-vs-PLL-3}
\end{minipage}


\section{Effects of word frequency}

\subsection{Different word contexts} \label{app:frequency-contexts}

\noindent%
\begin{minipage}{\linewidth}
\makebox[\linewidth]{
  \includegraphics[width=\textwidth]{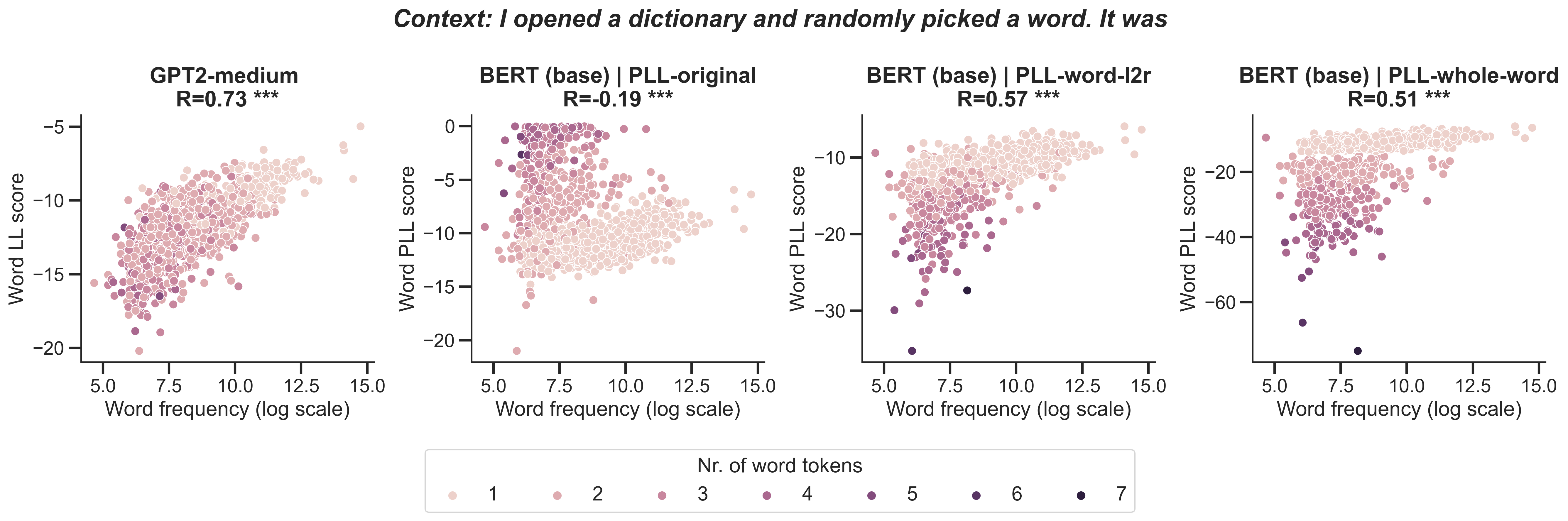}}
\captionof{figure}{Word frequency effects for {\tt bert-base-cased} on the EventsAdapt corpus. Word scores were retrieved with a neutral context: ``{\tt I opened a dictionary and randomly picked a word. It was} \_''.} \label{fig:word-freq-vs-PLL-2}
\end{minipage}

\noindent%
\begin{minipage}{\linewidth}
\makebox[\linewidth]{
  \includegraphics[width=\textwidth]{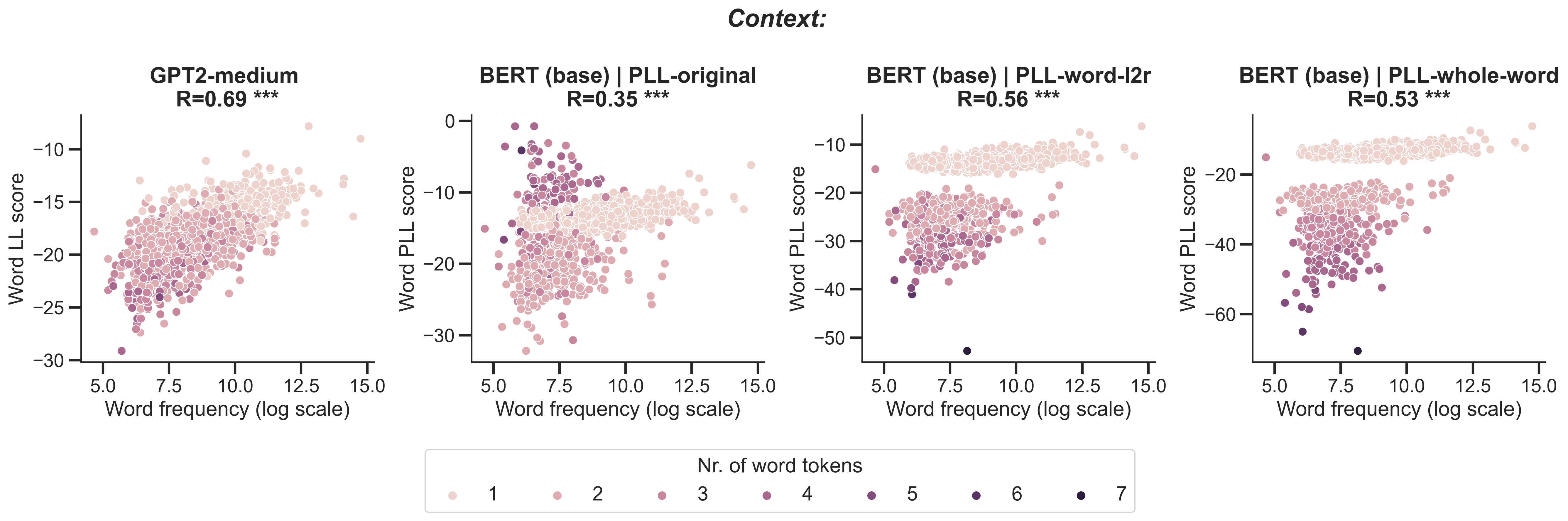}}
\captionof{figure}{Word frequency effects for {\tt bert-base-cased} on the EventsAdapt corpus. Word scores were retrieved without supporting context.}\label{fig:word-freq-vs-PLL-1}
\end{minipage}

\subsection{Different datasets}

\noindent%
\begin{minipage}{\linewidth}
\makebox[\linewidth]{
  \includegraphics[width=\textwidth]{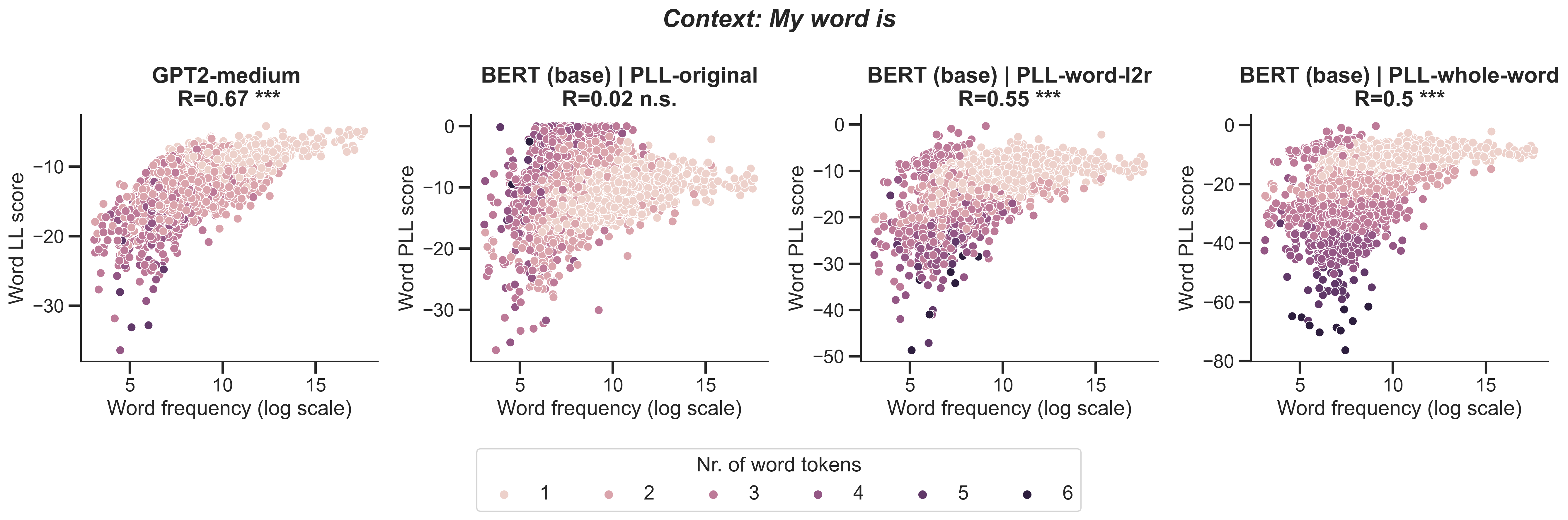}}
\captionof{figure}{Word frequency effects for {\tt bert-base-cased} on the LibriSpeech corpus. Word scores were retrieved with a neutral context: ``{\tt My word is} \_''.}\label{fig:word-freq-vs-PLL-3}
\end{minipage}

\noindent%
\begin{minipage}{\linewidth}
\makebox[\linewidth]{
  \includegraphics[width=\textwidth]{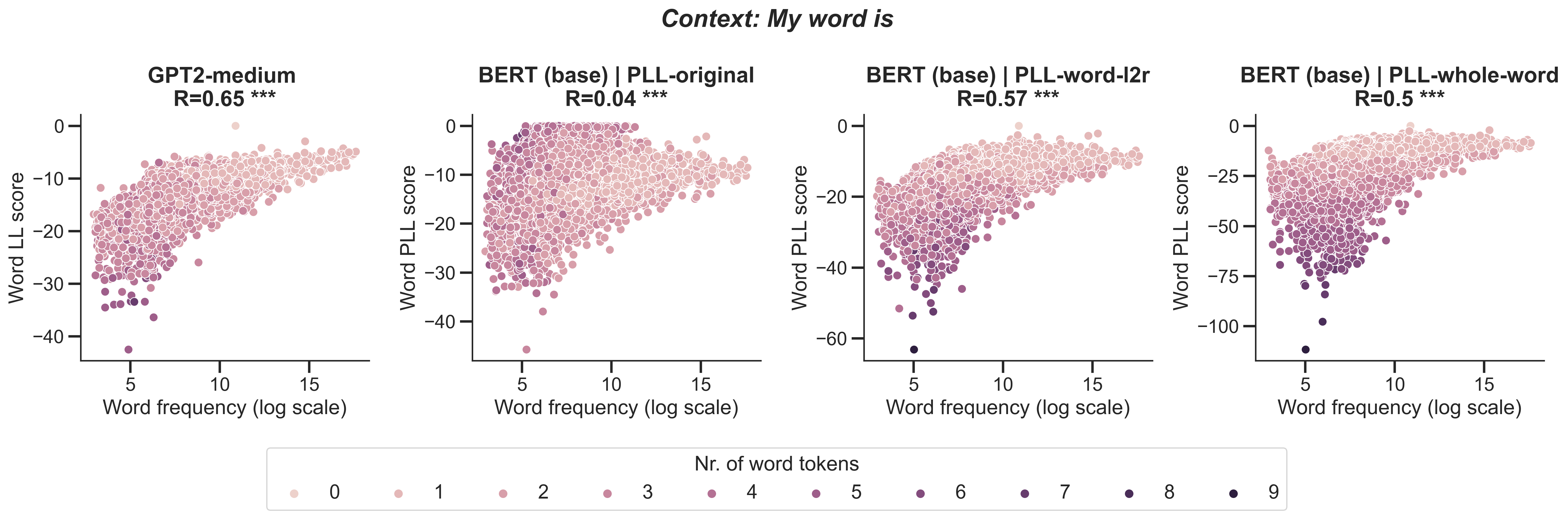}}
\captionof{figure}{Word frequency effects for {\tt bert-base-cased} on the Brown corpus. Word scores were retrieved with a neutral context: ``{\tt My word is} \_''.}\label{fig:word-freq-vs-PLL-4}
\end{minipage}


\section{Correlation with unidirectional models} \label{app:gpt-correlation}

\subsection{Larger LLM versions}

\noindent%
\begin{minipage}{\linewidth}
\makebox[\linewidth]{
  \includegraphics[width=\textwidth]{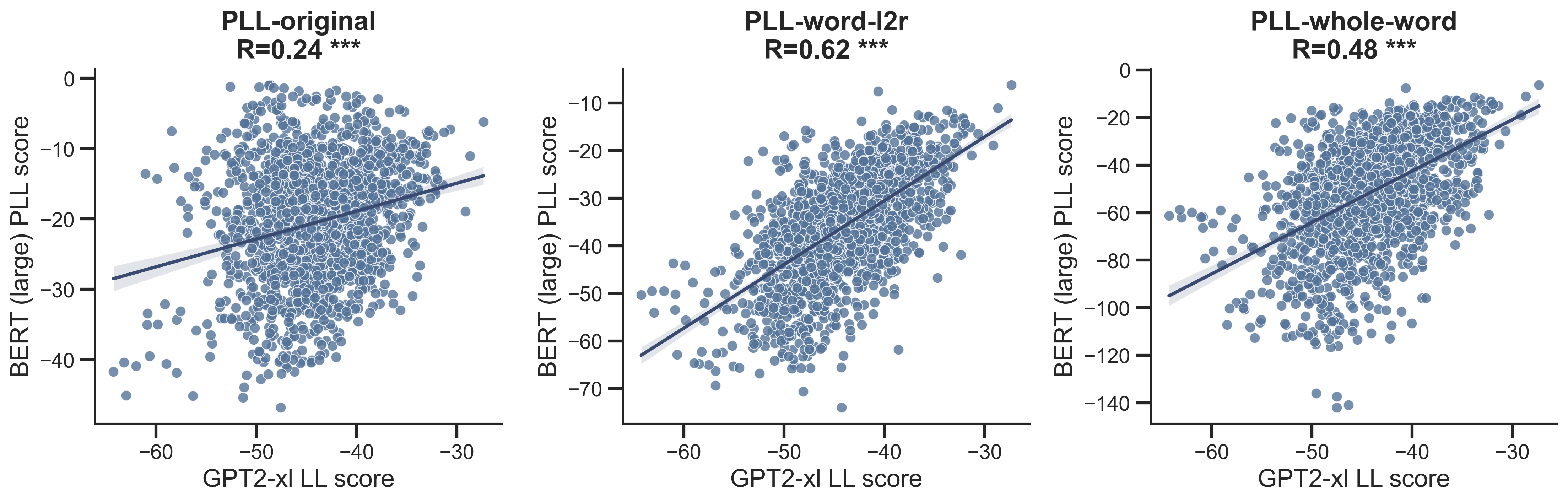}}
\captionof{figure}{Correlation between {\tt bert-large-cased} and {\tt gpt2-xl} scores on the EventsAdapt corpus.}\label{fig:gpt-vs-bert-large}
\end{minipage}

\subsection{Larger datasets}

\noindent%
\begin{minipage}{\linewidth}%
\makebox[\linewidth]{%
  \includegraphics[width=\textwidth]{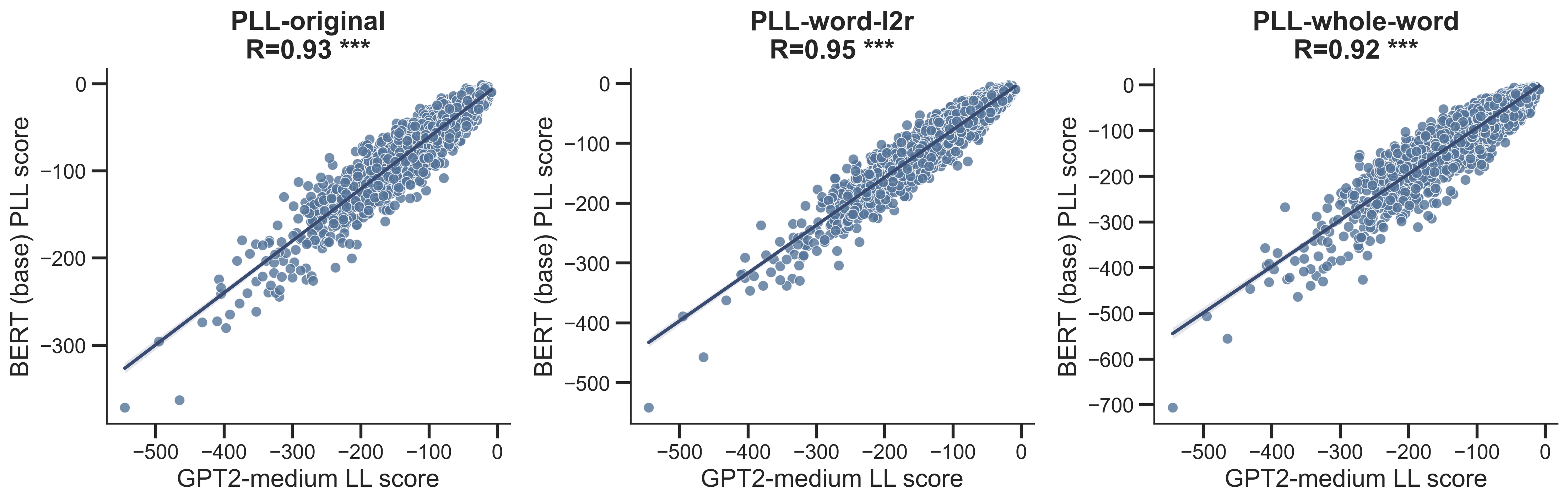}}
\captionof{figure}{Correlation between {\tt bert-base-cased} and {\tt gpt2-medium} scores on the LibriSpeech corpus.}\label{fig:gpt-vs-bert-librispeech}%
\end{minipage}

\noindent%
\begin{minipage}{\linewidth}%
\makebox[\linewidth]{%
  \includegraphics[width=\textwidth]{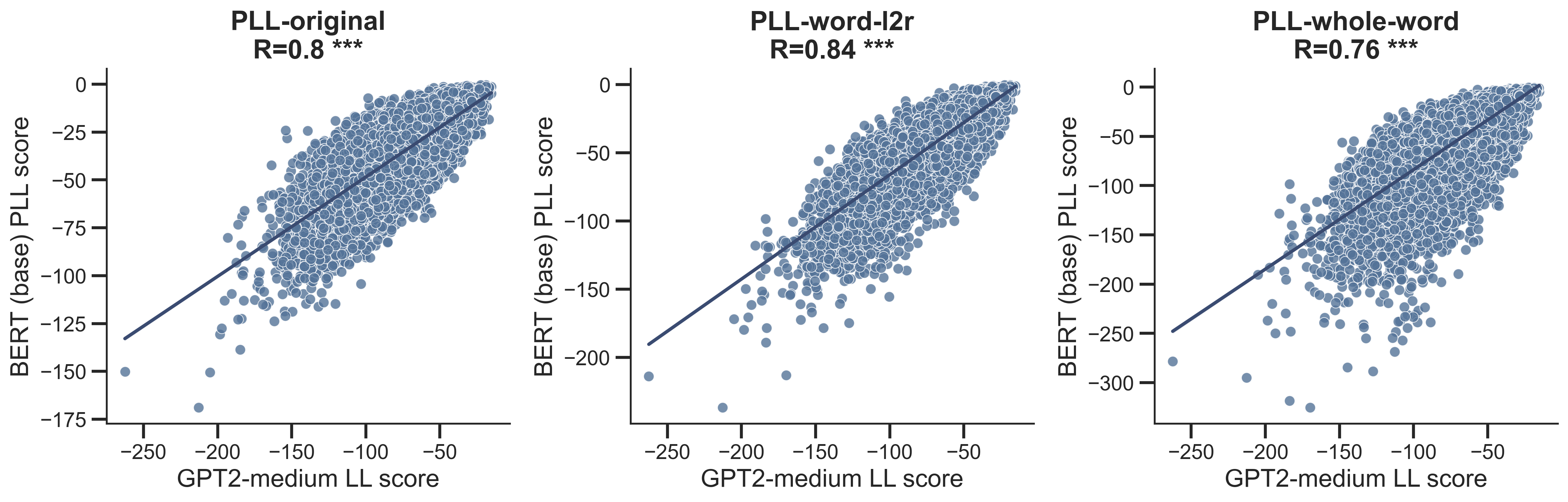}}
\captionof{figure}{Correlation between {\tt bert-base-cased} and {\tt gpt2-medium} scores on the Brown corpus.}\label{fig:gpt-vs-bert-brown}%
\end{minipage}


\section{Whole-sentence left-to-right token masking} \label{app:sentence-l2r}

Here, we report results for the scoring algorithm that masks the target token, $s_t$, and all sentence tokens to its right in a sentence $S$ with $n$ tokens ({\tt PLL-sentence-l2r}). As in autoregressive language models, target token inference is thus conditioned solely on the token's leftward context:  $P_{\mathrm{MLM}}(s_{t}~|~S_{<t})$. The final sentence score is obtained as the sum of the log probabilities of each sentence token given its context:
\vspace*{-1em}

\begin{equation}
    \mathrm{PLL}_{\mathrm{sent}}(S) := \sum_{t=1}^{n}  \mathrm{log}~P_{\mathrm{MLM}}(s_{t}~|~S_{<t})
\end{equation}

Overall, the {\tt PLL-sentence-l2r} metric satisfies the metric desiderata better than the {\tt PLL-original} metric but worse than {\tt PLL-word-l2r}. In addition, it is inferior to other metrics on the BLiMP evaluation benchmark (Appendix \ref{app:blimp}), in line with previous reports of subpar generation quality \cite{wang-cho-2019-bert}. \newline 
\vspace*{0.1em}

\noindent%
\begin{minipage}{\linewidth}%
\makebox[\linewidth]{%
  \includegraphics[width=0.6\textwidth]{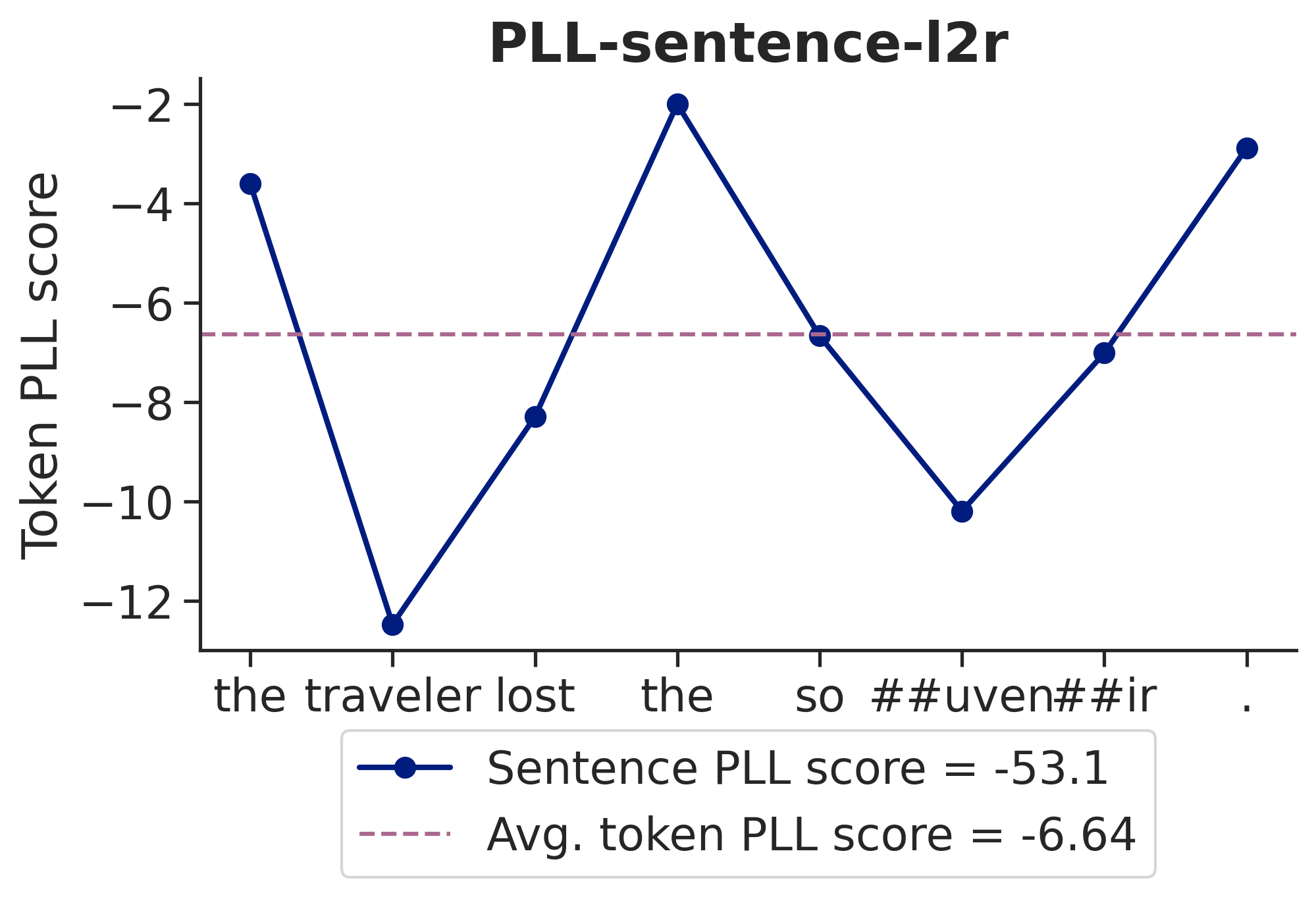}}
\captionof{figure}{Scores for the motivating example computed with {\tt PLL-sentence-l2r} ({\tt bert-base-cased}).}\label{fig:motivation-globall2r}%
\end{minipage}

\noindent%
\begin{minipage}{\linewidth}%
\makebox[\linewidth]{%
  \includegraphics[width=\textwidth]{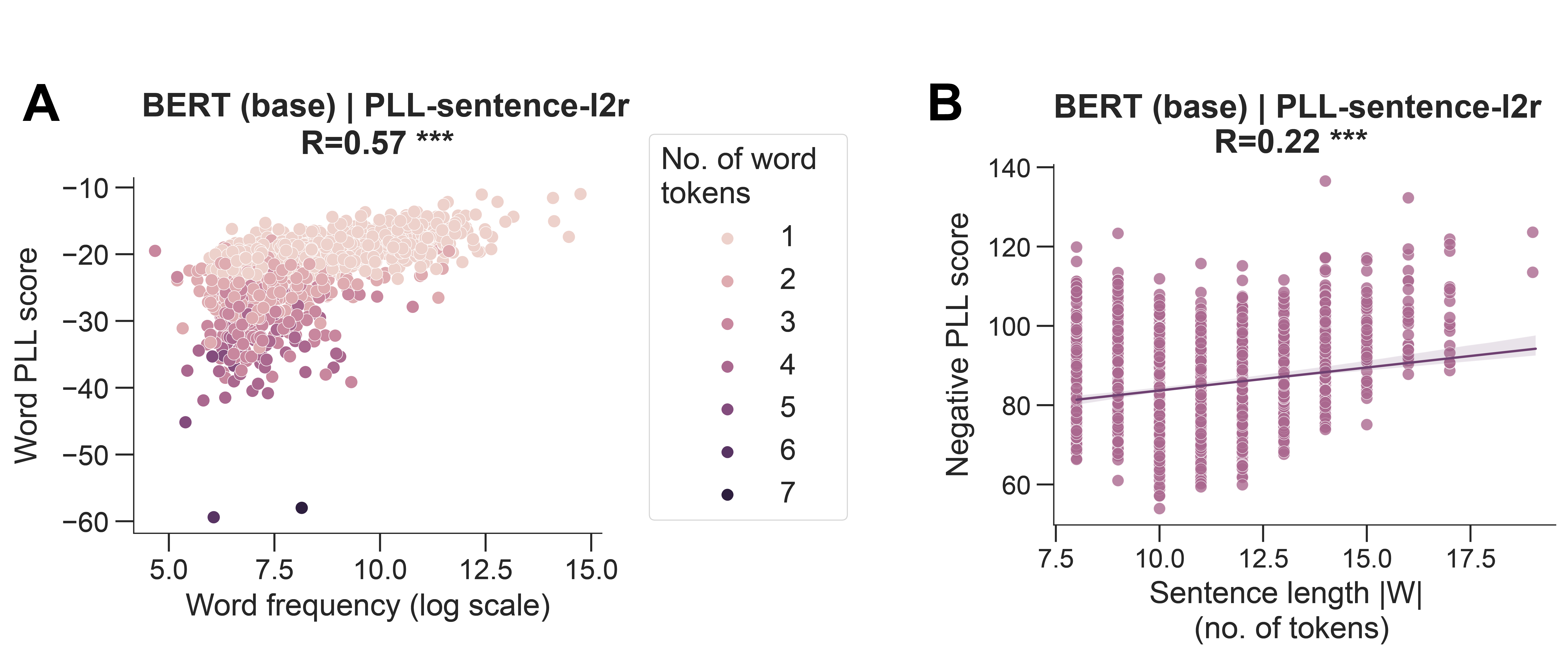}}
\captionof{figure}{Word frequency \textbf{(A)} and sentence length \textbf{(B)} effects for scores computed with {\tt PLL-sentence-l2r} on the EventsAdapt corpus ({\tt bert-base-cased})}.\label{fig:wordfreq_sentlength-globall2r}%
\end{minipage}
\vspace*{0.1em}

\noindent%
\begin{minipage}{\linewidth}%
\makebox[\linewidth]{%
  \includegraphics[width=0.4\textwidth]{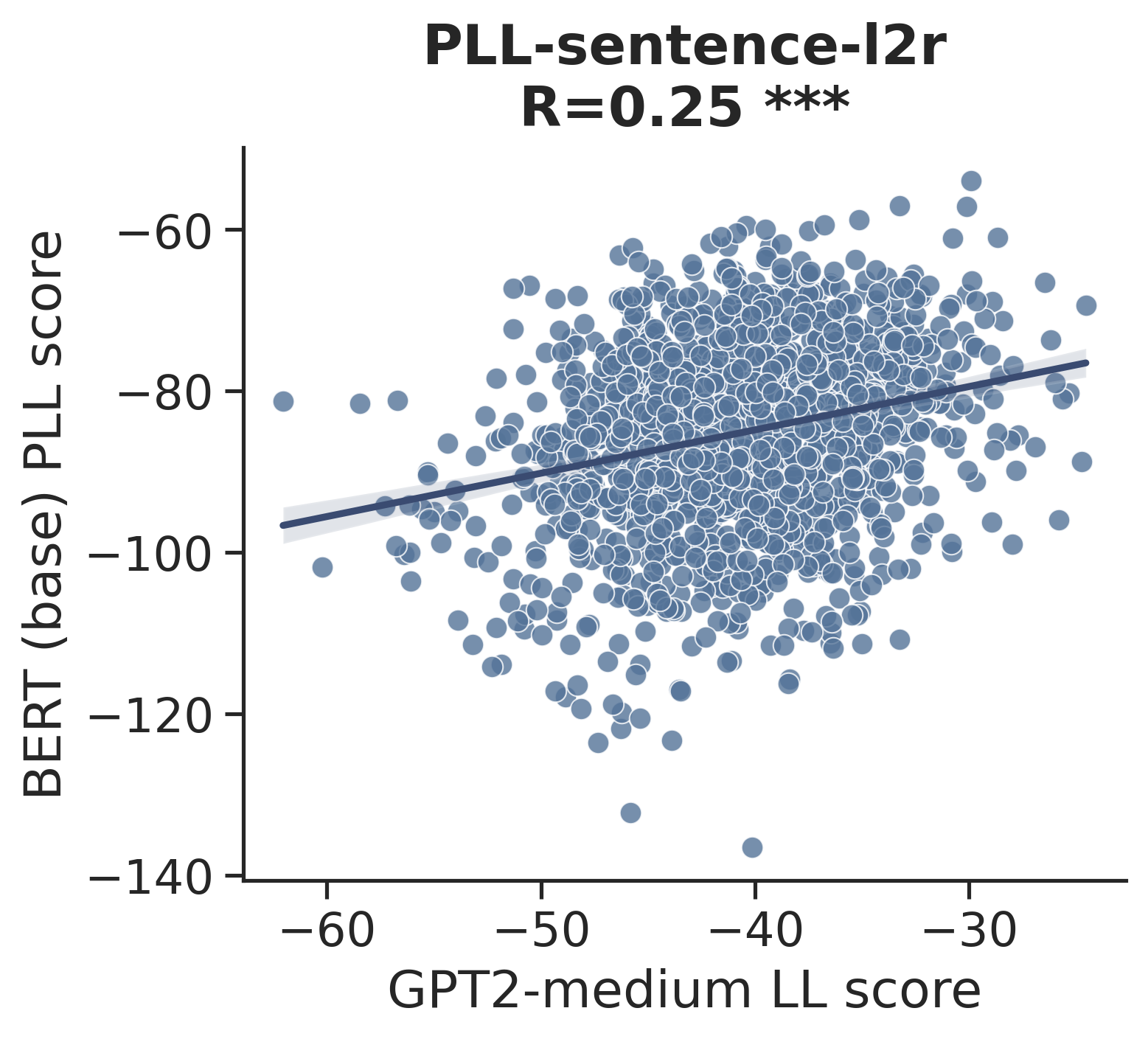}}
\captionof{figure}{Correlation between {\tt bert-base-cased} and {\tt gpt2-medium} scores computed with {\tt PLL-sentence-l2r} on the EventsAdapt corpus.}\label{fig:gpt-vs-bert-globall2r}%
\end{minipage}

\section{Detailed BLiMP benchmark results} \label{app:blimp} 

Table \ref{table:BLiMP-appendix} shows results for each sentence suite within the BLiMP benchmark (in addition to the overall scores reported in the main text). All models shown in Tables \ref{table:BLiMP} and \ref{table:BLiMP-appendix} are cased models. {\tt PLL-original} scores replicate those reported in \citet{salazar2020masked}.

\begin{table*}[t]
  \scriptsize
  \centering
\begin{tabular}{ll>{\columncolor[gray]{0.95}}ccccccccccccc}
{} & {} &  \rot{ \textbf{Overall}} &  \rot{ ANA. AGR} &  \rot{ ARG STR.} &  \rot{ BINDING} &  \rot{ CTRL. RAIS.} &  \rot{ D-N AGR} &  \rot{ ELLIPSIS} &  \rot{ FILLER GAP} &  \rot{ IRREGULAR} &  \rot{ ISLAND} &   \rot{ NPI} &  \rot{ QUANTIFIERS} &  \rot{ S-V AGR} \\
\toprule
\multirow{4}{*}{BERT (base)} & PLL-original       &     84.2 &           97.0 &              80.0 &              \textbf{82.3} &         79.6 &              97.6 &              89.4 &                  \textbf{83.1} &       96.5 &    73.6 &  84.7 &         \textbf{71.2} &     \textbf{92.4} \\
 & PLL-word-l2r       &     \textbf{84.7} &  \textbf{97.1} &      \textbf{81.0} &     \textbf{82.3} &         \textbf{81.9} &     \textbf{98.4} &      \textbf{89.6} &        83.0 &       96.5 &    \textbf{75.0} &  \textbf{85.0} &         69.8 &     92.1 \\
 & PLL-whole-word     &     83.1 &           96.6 &              76.5 &                  81.5 &               80.5 &             96.9 &              87.1 &                  82.5 &       \textbf{97.1} &    74.9 &  83.8 &         69.2 &     88.5 \\
 & PLL-sentence-l2r     &     58.7 &      80.3 &      63.0 &     68.3 &         53.5 &     82.1 &      68.3 &        47.8 &       47.3 &    56.5 &  38.9 &         51.6 &     50.7 \\
\midrule
\multirow{4}{*}{BERT (large)} & PLL-original      &     84.8 &           \textbf{97.2} &      \textbf{80.7} &     \textbf{82.0} &         82.7 &     97.6 &      \textbf{86.4} &        \textbf{84.3} &       \textbf{92.8} &    77.0 &  83.4 &         \textbf{72.8} &     \textbf{91.9} \\
 & PLL-word-l2r      &     \textbf{85.0} &  96.8 &      80.6 &     81.9 &         \textbf{84.8} &     \textbf{97.8} &      85.8 &        84.0 &       92.0 &    \textbf{78.8} &  \textbf{83.6} &         71.7 &     91.2 \\
 & PLL-whole-word    &     82.6 &           96.6 &      75.7 &     79.9 &         81.4 &     95.2 &      83.6 &        83.3 &       90.1 &    78.7 &  81.5 &         70.4 &     86.7 \\
 & PLL-sentence-l2r    &     59.8 &      61.5 &      63.0 &     71.3 &         60.5 &     71.8 &      58.3 &        58.5 &       63.0 &    50.2 &  42.8 &         51.9 &     63.0 \\
\midrule
\multirow{4}{*}{RoBERTa (base)} & PLL-original    &     85.4 &           97.3 &      83.5 &     77.8 &         81.9 &     97.0 &      91.4 &        \textbf{90.1} &       \textbf{96.2} &    80.7 &  81.0 &         \textbf{69.8} &     91.9 \\
 & PLL-word-l2r    &     \textbf{86.7} &  \textbf{97.8} &      \textbf{84.8} &     \textbf{78.7} &         \textbf{84.9} &     \textbf{98.3} &      \textbf{91.6} &        90.0 &       95.4 &    \textbf{81.0} &  84.4 &         69.7 &     \textbf{94.0} \\
 & PLL-whole-word  &     85.4 &           97.6 &      80.9 &     76.6 &         85.2 &     96.6 &      \textbf{91.6} &        90.0 &       95.6 &    80.2 &  \textbf{84.7} &         69.6 &     91.1 \\
 & PLL-sentence-l2r  &     79.3 &      97.0 &      79.9 &     71.2 &         78.4 &     95.0 &      84.8 &        82.6 &       85.0 &    68.2 &  80.6 &         58.4 &     81.6 \\
\midrule
\multirow{4}{*}{RoBERTa (large)} & PLL-original   &     86.5 &           97.8 &      84.6 &     79.1 &         84.1 &     96.8 &      \textbf{90.8} &        88.9 &                   \textbf{96.8} &  \textbf{83.4} &  85.5 &                   70.2 &    91.4 \\
 & PLL-word-l2r   &     \textbf{87.5} &  98.0 &      \textbf{85.0} &     \textbf{80.0} &         86.8 &     \textbf{98.3} &      90.4 &        \textbf{89.1} &                  95.7 &  \textbf{83.4} &  \textbf{88.0} &         \textbf{70.3} &     \textbf{93.2} \\
 & PLL-whole-word &     85.9 &           98.2 &      80.2 &     78.0 &         \textbf{87.1} &     96.0 &      90.1 &        88.9 &       95.6 &    82.2 &  \textbf{88.0} &         69.8 &     89.7 \\
 & PLL-sentence-l2r &     80.4 &      \textbf{98.8} &      82.5 &     71.8 &         80.4 &     95.1 &      82.0 &        80.8 &       91.6 &    73.0 &  76.6 &         57.8 &     86.0 \\
\midrule
 \emph{Human} &                     &     \emph{88.6} &           \emph{97.5} &     \emph{90.0} &     \emph{87.3} &     \emph{83.9} &     \emph{92.2} &     \emph{85.0} &     \emph{86.9} &     \emph{97.0} &     \emph{84.9} &     \emph{88.1} &     \emph{86.6} &     \emph{90.9} \\
\bottomrule
\end{tabular}
\caption{\normalsize Unsupervised performance (forced choice accuracy) on all BLiMP benchmark paradigms, using the original and adjusted PLL sentence scoring methods. {\tt PLL-original} scores replicate those reported in \citet{salazar2020masked}. Human scores are taken from \citet{warstadt2020blimp}.}
\label{table:BLiMP-appendix}
\end{table*}

\end{document}